\documentclass[10pt,twocolumn,letterpaper]{article}
\pdfoutput=1
\usepackage{cvpr}
\usepackage{times}
\usepackage{epsfig}
\usepackage{graphicx}
\usepackage{amsmath}
\usepackage{amssymb}
\usepackage{multicol, blindtext}

\usepackage[breaklinks=true,bookmarks=false]{hyperref}

\cvprfinalcopy 

\def\cvprPaperID{****} 

\begin{document}

\title{Incremental Learning in Person Re-Identification}

\author{Prajjwal Bhargava\\
{\tt\small prajjwalin@protonmail.com}
}

\maketitle

\begin{abstract}
   Person Re-Identification is still a challenging task in Computer Vision due to a variety of reasons. On the other side, Incremental Learning is still an issue since deep learning models tend to face the problem of over catastrophic forgetting when trained on subsequent tasks. In this paper, we propose a model that can be used for multiple tasks in Person Re-Identification, provide state-of-the-art results on a variety of tasks and still achieve considerable accuracy subsequently. We evaluated our model on two datasets Market 1501 and Duke MTMC. Extensive experiments show that this method can achieve Incremental Learning in Person ReID efficiently as well as for other tasks in computer vision as well. The code for this work can be found \href{https://github.com/prajjwal1/person-reid-incremental}{here} 
\end{abstract}

\section{Introduction}

Deep neural networks have revolutionized the field of computer vision. In recent years, a lot of work has been done in Person Re-Identification, we've seen considerable progress but are faced with a lot of challenges in terms of getting accurate predictions in real-life instances. It plays an important role in many areas, surveillance being one of them. In some sense, it can be compared to other prominent tasks in computer vision like Image classification, where a lot of progress has been made. Moreover, there has been a growing demand for deep learning models that incur the low computational cost. Deployment of such models can be cumbersome and may not prove to be much efficient especially if the same task can be carried out with a lesser number of parameters. Given a set of images of a person taken from different angles from a different camera, our model is required to generate a higher prediction if those images are of the same person and vice versa. The problem is composed of multiple reasons some of which may include background clutter, illumination conditions, occlusion, body pose, the orientation of cameras. Numerous methods have been proposed to address some of these issues. So far the models that have been proposed in Person ReID are good in doing well in a particular dataset but when tested on a quite similar dataset, they struggle to get accurate predictions. Unlike other tasks such as Image Classification or Object Detection, we are required to have our model perform well on a large number of classes and all these images are not as much distinctive as other objects do which makes it difficult for the neural network to generate accurate predictions. We devise a new method that can be used to create robust Person ReID systems at a lower computational cost that can not only perform well on one task but if trained properly using our techniques, it can be well adapted to other tasks as well.

\begin{figure*}
\includegraphics[width=\textwidth,height=4cm]{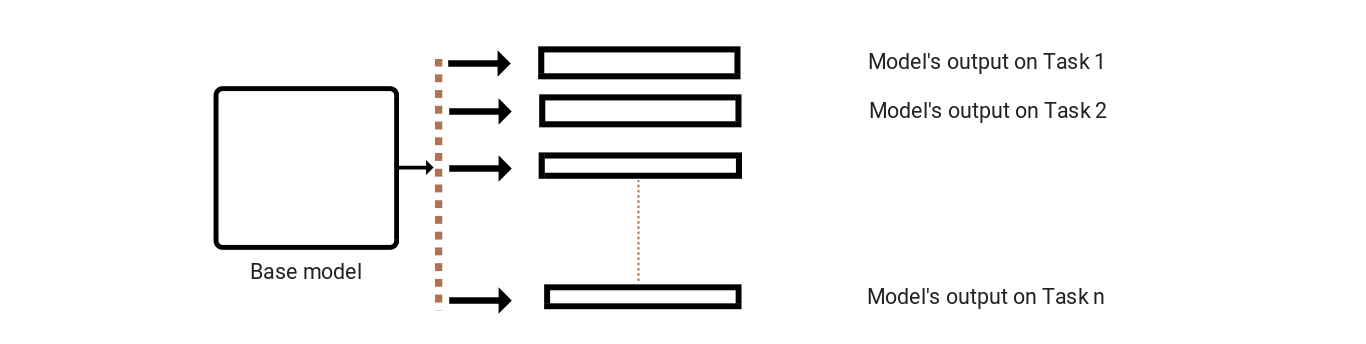}
\caption{Proposed architecture}
\end{figure*}
\section{Related work}

For Incremental Learning, much research work has been carried out. Our work is slightly inspired by Learning without forgetting\cite{li2016learning}, which was used for classification purposes. They made use of CIFAR10 and SVHN as the two tasks and then achieved considerable performance. Other closely associated work which builds upon it is SeNA-CNN\cite{zacarias2018a}, wherein they made the architecture a little more complex by introducing more layers in different pipelines instead of just dealing with fully connected layers. Our work is the first one to our knowledge that tries to tackle the problem of Incremental Learning in Person Re-Identification, unlike image classification where we have a relatively lesser number of classes, the number is way more, and this increases the difficulty level for generating accurate predictions. In defense\cite{DBLP:journals/corr/HermansBL17} made use of Triplet Loss to show that it can be used to perform end to end deep metric learning. Some work that has been carried out in this incremental learning space also makes use of Distillation\cite{44873} loss wherein you train a smaller network to produce close predictions to cumbersome models. But to carry out this task, we are also required to train our cumbersome model first to be able to train the smaller model which is again a big task. Our proposed method doesn't rely on multiple models or older data that has been used to train the network on earlier task, rather we have multiple heads inside one model which aims to resolve this issue.

\section{Our proposed method}
We propose a new architecture that's relatively simple as compared to other proposed methods for achieving Incremental Learning along with a few techniques that make convergence faster and increases the model's accuracy. 

\subsection{Overall architecture}
We use a ResNet50\cite{DBLP:journals/corr/HeZRS15} which has been pretrained on ImageNet\cite{DBLP:journals/corr/RussakovskyDSKSMHKKBBF14}. We remove the last two layers i.e Fully connected layer and Average Pooling layer. We then introduce multiple heads. The main goal behind keeping ResNet is to perform the task of feature extraction effectively, it acts as a base model that contains global common features extracted from the data. This can also be done by any other classification network as well. This is based on the assumption that the tasks would have some sort of similar characteristics that might be common amongst all tasks on which the network is going to be trained. We then use these heads to generate task-specific predictions. These pipelines can be modified in accordance with use cases to better adapt to a given task. In our case, since the two tasks were similar, we decided to keep them identical. In this context, the use of heads and pipeline is interchangeable.

\subsection{Multiple Pipelines}
We introduced two pipelines after the base model, one is meant to work on Market1501\cite{zheng2015scalable} and the other works on DukeMTMCC\cite{DBLP:journals/corr/abs-1712-09531}. Each pipeline consists of two convolutional blocks followed by a Fully connected layer. Each convolutional block consists of convolutional layer which takes in \textit{n} input channels with kernel size 1,stride 1 and outputs \textit{n/2} channels. This is followed by Batch Normalization\cite{DBLP:journals/corr/IoffeS15} and usage of Leaky ReLU\cite{DBLP:journals/corr/abs-1803-08375} activation. Another block takes in \textit{n} input channels and outputs \textit{n} channels with kernel size 3 and stride kept to 1. So in this process dimensionality is not changed. Later the input is then fed to a Fully connected layer, to generate the prediction vector depending upon the number of classes we require. Residual connections in case of several layers are bound to help and would also reduce the need for more number of parameters by a greater amount.


\subsection{Optimizer}
We initially tried Adam\cite{DBLP:journals/corr/KingmaB14}, which gave an accuracy of 74\% on Rank 1 on Market1501 dataset followed by weight decay. We then tried SGD with Cyclical Learning Rate (CLR)\cite{DBLP:journals/corr/Smith15a} scheduler which helped us achieve much higher accuracy. We saw an increment of more than 10\% on Rank 1 on Market 1501 to reach 89.3\%. We use the triangular variant with default values as suggested. We restricted our batch size to 32 as it provided the best results. Keeping a higher batch size would lead to less frequent weight updates. Since the learning rate becomes variable with CLR, it can take advantage of its behavior of making learning rate variable wherever necessary in a more effective manner as our experiments have shown. 

\begin{center}
 \begin{tabular}{||c c c ||} 
 \hline
 No. & Batch Size & Rank1 (Market1501) \\ [0.5ex] 
 \hline\hline
 1 & 32 & 89.3\% \\ 
 \hline
 2 & 64 & 79.2\% \\    [1ex] 
 \hline
\end{tabular}
\end{center}

\begin{figure*}
\includegraphics[width=\textwidth,height=4cm]{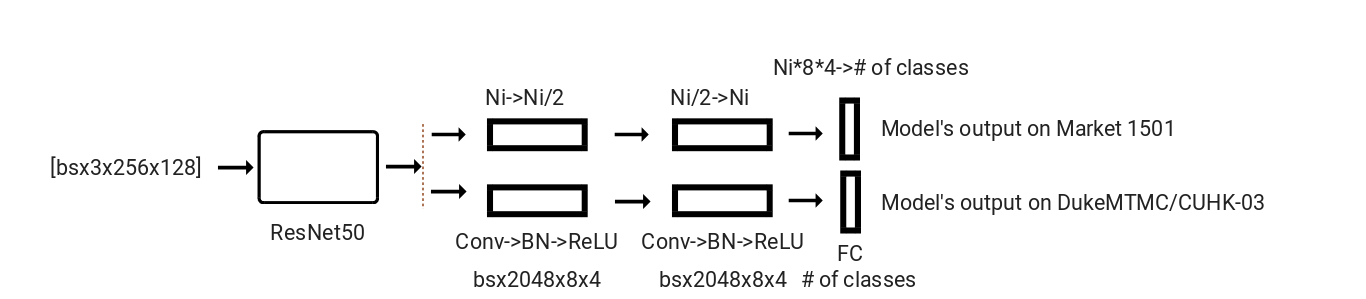}
\caption{Our architecture}
\end{figure*}

\subsection{Using Covariance loss for contrastive feature learning}

To deal with over catastrophic forgetting, We are proposing a new addition to our loss function whose main aim is to make positive targets (images of the same person, taken with a different camera) closer and negative targets (images of different person) far away in embedding space. We take feature maps that we get from the second convolution block from both the pipelines during the second phase. This is going to optimize embedding space such that data points with the same identity are closer to each other than those with different identities. We are required to take feature maps of positive targets and negative targets, we then have to perform the following operation:
\begin{equation}
A = \lambda*(\alpha*\sum((P_{i+1}-P_i)-(N_{i+1}-N_i))-\beta)
\end{equation}
where P and N denote features from positive and negative targets respectively and \textit{i} corresponding indexes.
$\lambda $, $\alpha$ and $\beta$ are three hyperparameters. Finding the most optimum value for these hyperparameters is an exhaustive process since we are required to let our model train for a considerable amount of epochs. A higher value of $\lambda$ introduces fluctuations in overall loss function since the value changes rapidly and may cause instability. We found that these set of values worked best in our case 
\begin{center}
 \begin{tabular}{||c c||} 
 \hline
 Hyperparameter & Value \\ [0.5ex] 
 \hline\hline
 $\lambda$ & 1 \\ 
 \hline
 $\alpha$ & 1e-9 \\ 
 \hline
 $\beta$ & 0 \\ 
 \hline
\end{tabular}
\end{center}

To perform this type of task we can either keep track of positive targets and negative targets before feeding them to the model or we can create a mask that can indicate which feature maps to choose. We use the second approach. Mask outputs a consecutive vector in pair, one for positive targets and the other for negative targets that indicate which feature maps to pick out of multiple maps. The dimension of feature maps returned from the intermediate layer of our network can vary depending upon the architecture of the base model. In our case, it is $ [batch\_size,2048,8,4] $. To operate subtraction, we flatten the feature map of both positive and negative targets followed by the addition of another axis giving us an embedding matrix of shape $[batch\_size,2048,1]$, which is then followed by a transpose operation.
This gives us a covariance embedding matrix whose sum of elements is going to give us an indicator to improve the model's predictions by restricting gradient flow in the base model.
\begin{equation}
X = \mathbf{A}*\mathbf{A}^\intercal\ 
\end{equation}

We tried different values of $\lambda$ to get the best accuracy possible. The value being computed by covariance loss introduces fluctuations on overall loss since cross-entropy reduces monotonically approximately in the initial and mid-course of the training, but that's not the case with covariance loss since weights of feature maps are changing rapidly relatively. So it's recommended to keep both the loss values in the same range. Therefore the value of $\lambda$ plays a great role in determining the overall performance of the model itself.
\begin{center}
 \begin{tabular}{||c c||} 
 \hline
 Value of $\lambda$ & Accuracy after 100 epochs(Rank 1) \\ [0.5ex] 
 \hline\hline
 0.7 & 83.3 \\ 
 \hline
 0.8 & 83.1 \\ 
 \hline
 0.9 & 83.1 \\ 
 \hline
 1 & \textbf{84.6} \\
 \hline
 1.3 & 82.9  \\ [1ex] 
 \hline
\end{tabular}
\end{center}

\subsection{Training methodology}
There are few ways to train these pipelines, we divided the training into two phases. In the first phase, only the base model along with the first pipeline was trained on Market 1501 and predictions were taken from the first pipeline itself. There can also be slight variation in the first phase, wherein some sections of the model can either be set as nontrainable or be used with discriminative learning rates. Other pipelines can be trained in a different manner (task-specific). In the second phase, we freeze the first pipeline and then train the base model along with the second pipeline on Duke MTMC and make predictions accordingly. A similar procedure can be repeated for n pipelines for n tasks.

\subsection{Objective Function}
Our loss function has two critical components now. We are using cross entropy as our classification loss along with our covariance loss. 
Cross Entropy is given as:
\begin{equation}
 H_{y'} (y) := - \sum_{i} y_{i}' \log (y_i) 
\end{equation}
where $y_i$ is the predicted probability value for class i and $ y_i'$ is the true probability for that class.
Our final loss is the sum of cross entropy and covariance loss.
\begin{equation}
A + H_{y'} (y)
\end{equation}

\section{Experiments}

\subsection{Datasets}

We used two datasets for this work. Although other datasets can be used, These datasets have the most number of images as compared to other prevalent datasets. Market1501 contains 32668 images of 1501 persons split into train/test sets of 12,936/19,732. It has bounding boxes from a person detector that have been selected based on their intersection-over-union overlap with manually annotated bounding boxes. Duke MTMC has 16,522 training images of 702 identities, 2,228 query images of the other 702 identities and 17,661 gallery images (702 ID + 408 distractor ID). The proposed model is bound to perform much better if it's trained on more data.
\begin{center}
 \begin{tabular}{||c c c ||} 
 \hline
 No. & Dataset & Num of identities \\ [0.5ex] 
 \hline\hline
 1 & Market 1501 & 751 \\ 
 \hline
 2 & Duke MMTC & 702 \\  [1ex] 
 \hline
\end{tabular}
\end{center}

\subsection {Ensembling}
Ensembling has often given improved results in various computer vision tasks. This often works well when predictions are being taken from multiple models. Here we tried ensembling amongst these two pipelines. The first phase was performed as usual. The second phase was trained with different ensembling combinations amongst different network modules. We noted that the model converged faster relatively and accuracy was saturated to a lower max value. Although it may prove to work better if a specific set of pipelines are used to solve a particular task.

\begin{table}[h!]
\centering
 \begin{tabular}{||c c||} 
 \hline
 Ensembling method & Rank1(epochs)\\ [0.5ex] 
 \hline
 Base model,second pipeline & 84.5(100),87.7(500)\\ 
 \hline
 Both pipelines & 84.1 (100)\\
 \hline
\end{tabular}
\end{table}

\subsection{Results}
Since our main goal is to bring generalization into our model and avoid over catastrophic forgetting, we first train the first pipeline, followed by the evaluation of predictions coming from the last FC layer of the first pipeline. Then we train the second pipeline followed by evaluation. In the last phase, we don't do any training and just evaluate it on the first task our model was made to perform. These results are reported after the model converged. 

\begin{table}[h!]
\centering
 \begin{tabular}{||c c c c c||} 
 \hline
 No. & Dataset& Rank1&Rank20&MaP\\ [0.5ex] 
 \hline\hline
 1 & Market 1501  & 89.3\% & 98.3\% & 71.8\% \\ 
 \hline
 2 & DukeMTMC & 80.0\%  & 93.7\%  & 60.2\% \\
 \hline
 3 & Market 1501 & 70.2\%  & 93.0\% & 41.2\% \\  [1ex] 
 \hline
\end{tabular}
\end{table}

We note state-of-the-art accuracies on both the tasks, and yet achieve considerable accuracy on the first task again. 

\section{Effectiveness of proposed method}

Our work indicates that we now have a simple method that can achieve state-of-the-art results when trained on Person Re-Identification tasks and yet achieve considerable accuracy on older tasks without losing much information and doesn't rely on older data after it has been used for training it. All of the learned information is distilled inside the model. This is a big step because we don't have access to older data in real-time instances and this would reduce the robustness of our model otherwise. Our architecture and discussed methods can be applied to other computer vision tasks as well. This method is bound to work with tasks that have fewer variations in the domain. For similar tasks, it seems to outperform other commonly used methods of training.

\section{Conclusion}
In this paper, we have shown that we can achieve incremental learning in Person ReID tasks with simpler methods yet achieving state-of-the-art results. We also propose a new novel loss that can be used to bring positive targets closer and negative targets farther in embedding space which results in improved performance for the desired task. We hope that our work would be built upon by Person ReID community to build better and robust incremental learning systems that can be further adapted to other domains as well thus increasing real-life usage of such systems. 

{\small
\bibliographystyle{ieee}
\bibliography{egbib}
}

\end{document}